\documentclass[final]{cvpr}

\usepackage{times}
\usepackage{epsfig}
\usepackage{graphicx}
\usepackage{amsmath}
\usepackage{amssymb}

\usepackage[ruled,vlined]{algorithm2e}
\usepackage{multirow}

\def\cvprPaperID{6124} 
\def\confYear{CVPR 2021}

\begin{document}

\title{Online Multiple Object Tracking with Cross-Task Synergy}




\author{Song Guo\thanks{Equal contributions}  $^1$ 
	\quad \quad Jingya  Wang\footnotemark[1]  $^{1,2}$ 
	\quad \quad Xinchao Wang$^{3,4}$ 
	\quad \quad Dacheng Tao$^1$ \\
$^1$The University of Sydney,\,\,
$^2$ShanghaiTech University,\\
$^3$National University of Singapore,\,\,
$^4$Stevens Institute of Technology\\
\tt\small sguo2908@uni.sydney.edu.au
\quad 
\tt\small wangjingya@shanghaitech.edu.cn
\\
\tt\small xinchao@nus.edu.sg
\quad 
\tt\small dacheng.tao@sydney.edu.au
}

\maketitle

\begin{abstract}
Modern online multiple object tracking (MOT) methods usually focus on two directions to improve tracking performance. One is to predict new positions in an incoming frame based on tracking information from previous frames, and the other is to enhance data association by generating more discriminative identity embeddings. Some works combined both directions within one framework but handled them as two individual tasks, thus gaining little mutual benefits. In this paper, we propose a novel unified model with synergy between position prediction and embedding association. The two tasks are linked by temporal-aware target attention and distractor attention, as well as identity-aware memory aggregation model. Specifically, the attention modules can make the prediction focus more on targets and less on distractors, therefore more reliable embeddings can be extracted accordingly for association. On the other hand, such reliable embeddings can boost identity-awareness through memory aggregation, hence strengthen attention modules and suppress drifts. In this way, the synergy between position prediction and embedding association is achieved, which leads to strong robustness to occlusions. Extensive experiments demonstrate the superiority of our proposed model over a wide range of existing methods on MOTChallenge benchmarks.
Our code and models are publicly available at \url{https://github.com/songguocode/TADAM}.

\end{abstract}

\section{Introduction}
The problem of multiple object tracking (MOT) has been studied for decades because of its broad applications such as robotics, surveillance, and autonomous driving. It aims to locate targets while maintain their identities to form trajectories across video frames.
Recent research in the area of MOT mostly follows the paradigm of tracking-by-detection, which divides the MOT problem into two separate steps. Object detections are obtained independently in each frame first, then linked across frames through data association to form trajectories, where identity embeddings are usually adopted to distinguish objects during association.
Such two-step procedure intuitively reveals two ways to improve the tracking performance. One is to augment detections, and the other is to enhance data association via embeddings.

\begin{figure}[t]
\begin{center}
    \includegraphics[width=1.0\linewidth]{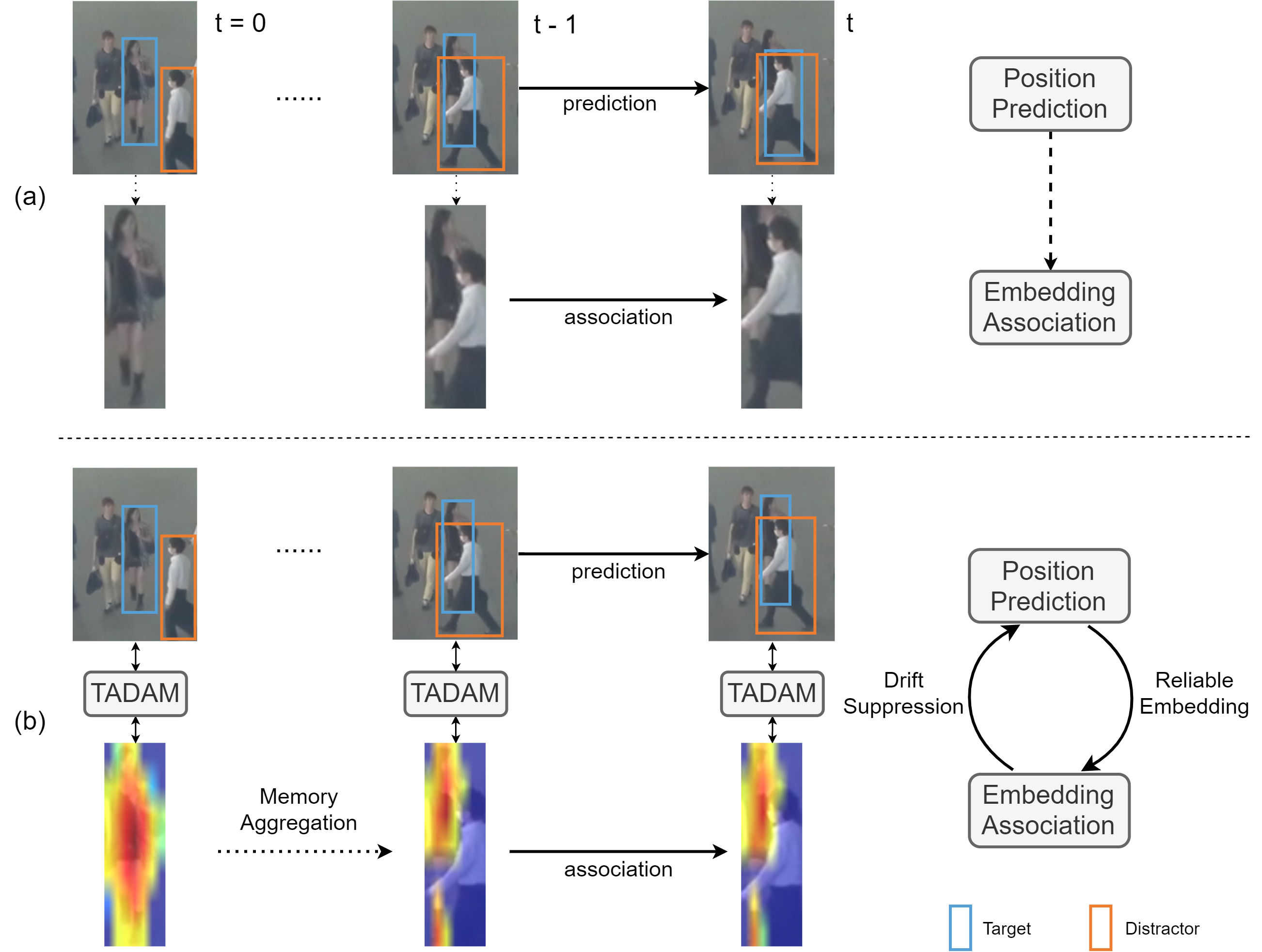}
\end{center}
   \caption{(a) Position prediction and embedding association in existing methods do not benefit each other during occlusion. The prediction of target position drifts and the extracted embeddings become noisy. (b) Our method brings synergy between the two tasks via proposed model to deal with occlusion.}
\label{fig:cover}
\end{figure}

Most existing online methods usually address only one of these two aspects for better tracking results, despite the fact that a common source of error, occlusions, affects both aspects. Unexpected occlusions often lead to miss detections due to overlapped objects, as well as increased difficulty for data association. 
Many online tracking approaches fill the gaps in detections during occlusions by predicting new positions of tracked targets, while a number of studies focus on generating more distinguishable embeddings to be associated throughout occlusions. 
Although a few recent works attempted to tackle both simultaneously, position prediction and embedding association were treated as two individual tasks.
How to make them benefit each other has not been well explored.

Common prediction methods seldom take interaction between objects into account, thus position prediction itself is not strong enough when dealing with occlusions. Making predictions under heavier occlusions often leads to drifted bounding boxes, where the predicted position of a target starts to follow a neighbouring object. The embedding extracted for association then deteriorates due to the wrongly predicted bounding box. This may lead to association errors that propagate over successive frames. Making predictions harm associating embeddings instead of helping in such cases. Meanwhile, improving embeddings alone only reduces errors at association stage, which does not help preventing position prediction errors at first hand. As such, there is no real synergy between the two tasks by treating them as two separate problems, as demonstrated in Fig. \ref{fig:cover}.

In this paper, we propose a unified model where position prediction and embedding association are jointly optimized with mutual benefits, boosting tracking performance by enhancing robustness to occlusions. To bring a real synergy, we make one task participate in the other's process. The two tasks are bridged by a link consisting of a target attention module and a distractor attention module, as well as a discriminative memory aggregation. Identity embeddings optimized for association are not only used in calculating affinity, but also applied to generate focus on a target as well as suppress drift through attention modules. In this way, the position prediction is equipped with identity-awareness and becomes sensitive to nearby objects, where more correct predictions can be performed under heavier occlusions without drifts. With better predictions during occlusions and attentions on the target, higher-quality embeddings can be extracted. Such more reliable embeddings then participate in the attention generation for better focus on the target. As a result, position prediction and embedding association are involved with each other, thus form a positive feedback loop with mutual benefits. The synergy is further amplified by an identity-aware memory aggregation, as richer holistic embeddings accumulated over time enable more robust attention generation. Consequently, tracking performance in complicated scenes with occlusions is boosted. We jointly optimize the position prediction, embedding association, and all proposed modules under a unified end-to-end model.
To the best of our knowledge, we are the first to achieve synergistic joint optimization on the two tasks.

The main contributions of this paper are listed as follows:
\begin{itemize}
    \item We propose a unified online MOT model that brings mutual benefits between position prediction and embedding association, thus achieving stronger robustness to occlusions.
    \item We apply temporal-aware target attention and distractor attention as well as an identity-aware memory aggregation to link the two tasks.
    \item Our tracker achieves state-of-the-art performance on MOTChallenge benchmarks with public detections.\
\end{itemize}

\section{Related Work}
\paragraph{Challenges in tracking-by-detection.}
The tracking-by-detection paradigm has been commonly adopted in most modern online multiple object trackers \cite{Bewley2016,Wang2014_ECCV,Wang2016_TPAMI,Lan2020_IJCV,Maksai2016_CVPR,Maksai2017_ICCV,Wojke2017, Wang2017, Chen2018, Zhu2018, Xu2019, Bergmann2019a, Xu2020a, Liu2020}. Off-the-shelf detectors like DPM \cite{Redmon2016}, Faster R-CNN \cite{Ren2015}, SDP \cite{Yang2016}, YOLO \cite{Redmon2016}, or SSD \cite{Liu2016} are first applied to discover objects in each incoming frame, then followed by data associations, where objects found in different frames are linked to form trajectories. Although earlier approaches like JPDA \cite{Fortmann1980, Rezatofighi2015}, MCMC-DA \cite{Oh2009}, and MHT \cite{Blackman2004, Zhai2010, Kim2015} evaluate all possible associations and directly form most probable trajectories in one step, they have been considered inefficient and not scalable in modern online complex MOT scenes. Methods adopting the tracking-by-detection paradigm face challenges in its two steps when tracking complicated scenarios with more occlusions.\
On the one hand, detections given by the detector become inaccurate or even missing due to occlusions. Such imperfection often gives rise to intermittent or fragmented tracklets and therefore degrades the tracking result. On the other hand, associating objects under complex scenes requires association measurement with stronger discriminability among objects with different identities. To this end, many online MOT methods aim to improve MOT performance by tackling either of these two issues.

\paragraph{Position prediction with visual cue.}
To fill gaps in detections, many works propose to infer locations of objects when they are not correctly given. In offline methods where all frames are provided and processed together, interpolations are performed to deduce intermediate positions once two object instances across multiple frames are confirmed to have the same identity \cite{Shen2018, Braso2019, Peng2020a, Hornakova2020,Wang_TIP_2017,Lan_TIP_2018,Ren_TIP_2021, He_TIP_2016, Shen_TC_2018}. However, such batch processing is not applicable in online methods where decisions must be made without access to data beyond the latest frame. As a result, online methods adopt prediction on target positions to deal with the gaps. Prediction can be made solely with motion models, such as a linear model like the Kalman Filter applied in \cite{Bewley2016, Wang2017} and non-linear model like LSTM used in \cite{Milan2016, Alahi2016, Fernando2018, Zhang2019}, but relying on motion only cannot achieve comparable performance with approaches utilizing visual cues for position prediction. For example, correlation filters have been applied to estimate new positions by finding the highest response in a new frame with visual features extracted from previous frames \cite{Wang2017, Feichtenhofer2017, Zhao2018c}. Single object tracking (SOT) trackers like ECO \cite{Danelljan2017}, SiamFC \cite{Bertinetto}, and SiamRPN \cite{Li2018} can be adopted in MOT by initiating one tracker for each target \cite{Zhu2018, Feng2019, Chu2019}. While they do eliminate some gaps in detections, such trackers lack the ability to differentiate objects of the same class, and are therefore vulnerable to occlusions by distractors. As frequency of occlusion is much higher and distractors are less distinguishable in MOT, special design is usually necessary to make SOT trackers to fit MOT framework. Bounding box regression in the second step of two-stage detectors like Faster R-CNN \cite{Ren2015} can be used as a predictor for new positions, by extracting features with previous bounding boxes passed from previous frame and infer displacements of boxes \cite{Bergmann2019a}. However, its power is still limited when dealing with heavier occlusions. To address the issue of occlusion in online position prediction, we propose to enforce stronger focus on targets and strengthen resistance to distractors.

\begin{figure*}[t]
\begin{center}
    \includegraphics[width=1.0\linewidth]{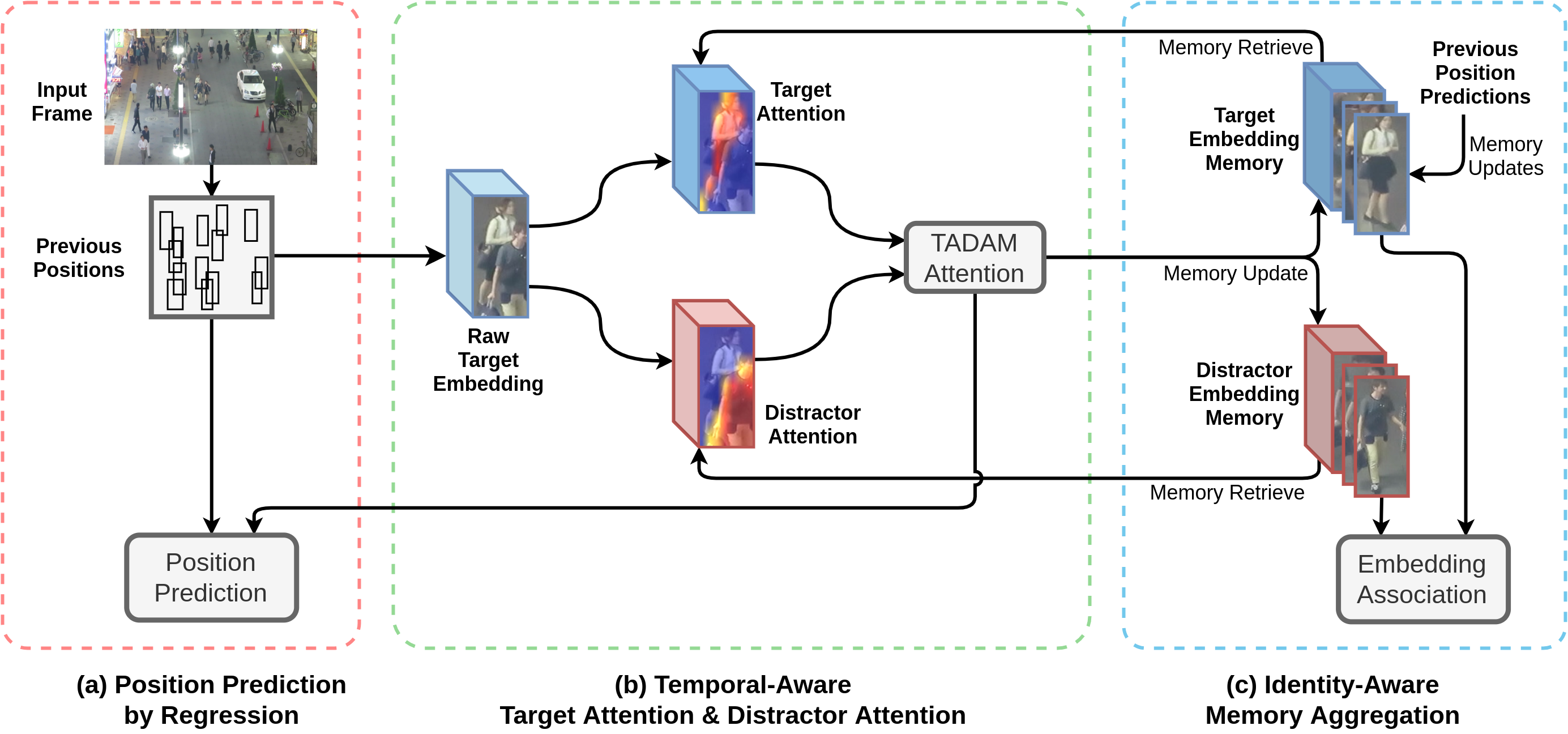}
\end{center}
   \caption{Model structure. The two tasks of position prediction and embedding association are bidirectionally linked by proposed modules to form synergy.}
\label{fig:structure}
\end{figure*}

\paragraph{Association with identity embeddings.}
Building a more reliable data association measurement is another direction to improve MOT performance. Earlier works link detections across frames with bounding box Intersection-over-Union (IoU) \cite{Bewley2016, Bochinski2017, Sheng2018a,Wang_CVIU_2014}, which is fast but not often inaccurate. Extracting an appearance embedding from each bounding box can establish a more discriminative association metric to distinguish objects of different identities. The identity embeddings can be used as main source of association \cite{Wojke2017, Chen2018}, or in conjunction with other features like motion feature \cite{Schulter2017, Wang2017, Sadeghian2017}. Such methods usually need a dedicated model trained with extra datasets, which incurs non-trivial cost in computation. More sophisticated association metrics can be built on fusion of identity embeddings with motion features \cite{Xiang2018, Chu2019, Yin2020}, or with fine-grained visual feature like body joints for pedestrian tracking \cite{Tang2017, Henschel2019}. Other approaches like layered tracking \cite{CongxiaDai2006, Bandouch2009, Chang2013} may also help identifying different targets. Accordingly, they all introduce much higher costs in the procedures of model designing and training.

More recently, UMA \cite{Yin2020} proposed to integrate embedding generation into its position prediction with a triplet structure, while DeepMOT \cite{Xu2020a} adopted an embedding head to produce identity embeddings simultaneously with regression-based position prediction. Such multi-task design lowers the cost for training association metric, but position prediction and association are treated as two individual tasks and their outcomes are not benefiting each other. While UMA \cite{Yin2020} designed a task-specific feature transformation to make the two tasks compatible under SOT framework, training an integrated embedding head in DeepMOT \cite{Xu2020a} has no impact on prediction results compared with having an externally trained embedding model. We show that the two tasks can work together in a more synergistic way that one of them participate in the improvement of another in tracking, which is essentially useful in complex scenes with occlusions.

\section{Proposed Method}

In this work, we propose a unified model that brings mutual benefit between position prediction and data association, so that robustness to occlusions is enhanced and tracking performance is boosted. To achieve this, we introduce temporal-aware target attention and distractor attention to form better focus on targets and suppress interference from distractors, as well as an identity-aware memory aggregation scheme for more robust attention generations. We name it by designed components as TADAM, where TA and DA refers to the target attention and the distractor attention, while M denotes the memory aggregation. All components are trained with the same data source within a unified model.
The overall framework of the proposed method is illustrated in Fig. \ref{fig:structure}.

\paragraph{Problem formulation.}
A tracked target object formed in a given sequence before frame \(t\) can be denoted as \(T^{ta}_{t-1}\), where its bounding box in frame \(t-1\) is described by \(B^{ta}_{t-1}\). A nearby distractor is described by \(T^{di}_{t-1}\) with its bounding box \(B^{di}_{t-1}\). 
\(F^{ta}_t\) represents the target's new position prediction feature extracted at frame \(t\) using \(B^{ta}_{t-1}\). While \(E^{ta}_t\) and \(E^{di}_t\) stands for similarly extracted latest identity embeddings of the target and its distractor respectively, their historical embedding references are given by \(E^{ta}_r\) and \(E^{di}_r\).

\subsection{Preliminary of position prediction by regression}
\label{sec:preliminary}
We adopt a regression-based position regression tracker \cite{Bergmann2019a} as a baseline, since it outperforms other prediction method with visual cues. It trains a two-stage Faster R-CNN detector with provided data, where an RPN is trained to generate coarse proposals boxes in the first stage, and a regression head together with a classification head are trained to refine boxes and deduce classes of objects inside boxes. During tracking, the first RPN stage is discarded, while the trained regression head is exploited to predict new position of a tracked target \(B^{ta}_t\) from prediction feature \(F^{ta}_t\) extracted at its previous location \(B^{ta}_{t-1}\), with the classification head giving the confidence for the prediction. Fig. \ref{fig:structure} (a), without connections with (b), illustrates the tracking procedure of this position prediction by regression. Embedding \(E^{ta}_t\) used for association is then obtained with an embedding extraction process. The power of position prediction mainly comes from inferring a tightly fitted bounding box with a given less accurate box, and it is trained with smooth L1 loss on displacements at four sides as adopted in \cite{Ren2015}.
Meanwhile, the classification head is learned with a cross-entropy loss between inferred classes and ground truth class annotations of input bounding boxes. Such position prediction approach gets rid of data association for targets being actively tracked, while matching through Hungarian algorithm is still necessary to search for potential reappearance of lost targets among new detections by comparing identity embeddings. Our proposed method aims to bring cross-task synergy on top of the position prediction design.

\subsection{Temporal-Aware Target Attention and Distractor Attention}
\label{sec:tada}
When a position prediction is performed on a target \(T^{ta}\) from frame \(t-1\) to \(t\), the new prediction feature \(F^{ta}_t\) in frame \(t\) is extracted with its previous bounding box \(B^{ta}_{t-1}\) and its new position can be predicted with \(F^{ta}_t\). However, when a distractor \(T^{di}\) is nearby and its bounding box, \(B^{di}_{t-1}\) has a large overlap with \(B^{ta}_{t-1}\), making a correct prediction becomes difficult. Suppose \(T^{ta}\) is occluded by \(T^{di}\), namely \(T^{di}\) is in front, then the predicted new bounding box of \(T^{ta}\) will tend to be closer to \(T^{di}\), as \(F^{ta}_t\) contains a large portion from \(F^{di}_{t}\) that actually belongs to \(T^{di}\). Continuing position prediction in such scenario will lead to a gradual drift of \(B^{ta}_t\) onto \(B^{di}_t\). To overcome such dominance, we introduce a target-attention (TA) module to augment regions in \(F^{ta}_t\) that belong to \(T^{ta}\) for better focus, as well as a distractor-attention (DA) module to suppress parts in \(F^{ta}_t\) belonging to \(T^{di}\) to reduce interference. The target attention is computed between the target's latest raw identity embedding \(E^{ta}_t\) and its historical aggregated embedding reference \(E^{ta}_r\), while the distractor attention is generated with \(E^{di}_t\) and the distractor's reference \(E^{di}_r\).
For simplicity, the distractor of a tracked target is selected as another nearby tracked target with largest IoU, where the one with highest overlapping is picked in case of multiple distractors.
With the two attentions applied on \(F^{ta}_t\) to obtain a refined prediction feature  \(\Tilde{F}^{ta}_t\), a better position prediction of \(B^{ta}_t\) can be performed. To further enhance the robustness of attention modules, a discriminative memory aggregation is designed to provide aggregated references of objects over time to make the attention modules discriminative and temporal-aware, which is covered in Sec. \ref{sec:aggregation}.

\paragraph{Discriminative aggregated non-local attention.}
To enhance or suppress regions in prediction feature \(F^{ta}_t\) for better prediction, we compute an attention projected from a reference embedding \(E^{ta}_r\) to the newly extracted raw embedding \(E^{ta}_t\). The dimension of embeddings is \(\mathcal{R}^{C \times H \times W}\), where \(C\) stands for channel, \(H\) for height, and \(W\) for width. \(E^{ta}_{t_i}\) and \(E^{ta}_{r_j}\) represents two points of dimension \(\mathcal{R}^{C}\) on \(E^{ta}_t\) and \(E^{ta}_r\) respectively, where \(i, j \in [1, HW]\) denotes two arbitrary spatial locations. With a discriminative aggregated target reference embedding \(E^{ta}_{r_j}\) as input, where the aggregation process is introduced in Sec. \ref{sec:aggregation}, an aggregated non-local target attention from historical memory reference \(E^{ta}_{r_j}\) onto \(E^{ta}_{t_i}\) can then be described as follows.
\begin{align}
  f(E^{ta}_{t_i}, E^{ta}_{r_j}) = \theta(E^{ta}_{t_i}) \phi(E^{ta}_{r_j}) \rho(E^{ta}_{r_j}),
  \label{eq:nonlocal target}
\end{align}
where \(\theta\) and \(\phi\) are convolution layers for computing correlation between the two, while \(\rho\) is another convolution layer to generate an representation of \(E^{ta}_{r_j}\) for output.

By serializing all location pairs \(j\) on \(E^{ta}_{r_j}\) and \(i\) on \(E^{ta}_{t_i}\), we obtain an overall non-local attention from the reference \(E^{ta}_r\) onto the new embedding \(E^{ta}_t\). Since \(E^{ta}_r\) is an aggregation with identity-aware memory, this process becomes a discriminative aggregated non-local attention between the target's historical references and its raw embedding obtained in a new frame.

Similarly, the discriminative aggregated non-local distractor attention from a distractor reference embedding \(E^{di}_{r_j}\) onto \(E^{ta}_{t_i}\) is given by equation below.
\begin{align}
  g(E^{ta}_{t_i}, E^{di}_{r_j}) = \theta(E^{ta}_{t_i}) \phi(E^{di}_{r_j}) \rho(E^{di}_{r_j}),
  \label{eq:nonlocal distractor}
\end{align}

While locations in the computed target attention with larger values indicate that those parts are more likely belong to the target, regions with higher response in distractor attention imply their greater probabilities of being parts of the distractor. We can then enhance the prediction with computation of refined prediction feature \(\Tilde{F}^{ta}_t\) given as follows.
\begin{align}
  \Tilde{F}^{ta}_t = F^{ta}_t \oplus w [f(E^{ta}_t, E^{ta}_r) \ominus g(E^{ta}_t, E^{di}_r)],
  \label{eq:refine}
\end{align}
where \(f\) and \(g\) stand for vectorized version of operations in Eq. \ref{eq:nonlocal target} and Eq. \ref{eq:nonlocal distractor}, while \(\oplus\) and \(\ominus\) denotes element-wise addition and subtraction. \(w\) denotes a weight to regulate attention output amount. \(E^{ta}_t\) is the newly extracted raw embedding in frame \(t\), while \(E^{ta}_r\) and \(E^{di}_r\) are the discriminative target reference and distractor reference retrieved from respective memories.

The combined temporal-aware target attention operation and distractor attention computation form TADAM attention for our model, as depicted in Fig. \ref{fig:structure} (b).
With \(\Tilde{F}^{ta}_t\) used for position prediction as mentioned in Sec. \ref{sec:preliminary}, the target's new position \(B^{ta}_t\) can then be predicted with more focus on itself and less interference from its distractor. Consequently, more correct predictions can be made under heavier occlusions, therefore allow collecting less noisy identity embeddings for data association and memory aggregation, which is discussed in Sec. \ref{sec:aggregation}.

\paragraph{Adaptive weight in target and distractor attention.}
While enhancement from TA and suppression from DA make position prediction more focused and less distracted, they are not necessarily useful for easy cases, especially on targets with little to no occlusion. Furthermore, applying a fixed weight \(w\) as in Eq. \ref{eq:refine} for TA output and DA output for all degrees of occlusions is suboptimal. Targets undergoing heavier occlusions should expect larger enhancement and stronger suppression, while those with little overlaps with neighbours should be left as is with no processing. To address this issue, an adaptive weight for target attention and distractor attention is designed as follows.
\begin{align}
  w = \frac{max(iou(B^{ta}_t, B^{di}_t) - o_{min}, 0)}{1 - o_{min}},
\end{align}
where \(iou(,)\) stands for computation of the IoU between two input boxes, \(B^{ta}_t\) represents a target's box and \(B^{di}_t\) refers to the box of its distractor. \(o_{min}\) gives the minimum level of overlapping for the weight to take effect, and \(max(,)\) outputs larger value of the two inputs. This value is computed per target-distractor pair.

The weight becomes non-zero and ranges between 0 to 1 when the computed IoU between a target and its distractor exceeds \(o_{min}\), otherwise assigned to 0. Output from target attention and distractor attention are adaptively regulated by this weight before participating further refinement.
Namely, target attention and distractor attention do not engage in easy cases with insignificant occlusions, while larger portion of their outcomes are used to handle more severe occlusions.

\subsection{Identity-Aware Memory Aggregation}
\label{sec:aggregation}
Reference embeddings of targets and those of distractors that participate in the target attention module and the distractor attention module respectively are described in Sec. \ref{sec:tada}. Although storing embeddings formed in the previous frame is an intuitive and accessible way to obtain references, embeddings obtained in such approach are usually noisy in more complex scenarios with heavier occlusions. To enhance the robustness of attention computations, we propose an identity-aware memory aggregation to accumulate more holistic references as inputs to the TA and DA attention modules, so that position predictions and embeddings can both be further boosted.

In each frame, target attention and distractor attention computed with newly obtained raw embedding \(E^{ta}_t\) of a target is used to produce refined prediction feature \(\Tilde{F}^{ta}_t\). Similarly, \(E^{ta}_t\) itself is also processed by the attentions to form refined embedding \(\Tilde{E}^{ta}_t\) for association and aggregation, like the generation of \(\Tilde{F}^{ta}_t\) in Eq. \ref{eq:refine}.
As the dimension of \(\Tilde{E}^{ta}_t\) to be aggregated is \(\mathcal{R}^{C \times H \times W}\), and the reference feature produced after aggregation needs spatial dimensions to be used for attention computation, we have to keep spatial information before and after aggregation. In addition, for a holistic aggregation, we expect the aggregation be able to automatically determine whether an input is worth updating, rather than naive accumulation that stores every input. To address these concerns, we design a discriminative memory module with convolutional gated recurrent unit (GRU), where matrix multiplications in GRU are replaced with convolutions. This builds a memory for temporal relation across frames as well as keep spatial dimensions.
The update of aggregated embedding memory is described as follows.
\begin{align}
    E_{r_t} = update(\Tilde{E}_t, E_{r_{t-1}}),
  \label{eq:memory}
\end{align}
where \(update()\) is the memory update function, \(E_{r_{t-1}}\) stands for previous state of the memory, and \(E_{r_{t}}\) refers to aggregated embedding updated with \(\Tilde{E}_t\) and \(E_{r_{t-1}}\). Note that this update process applies to both targets and distractors, where \(E\) is replaced by \(E^{ta}\) for target embedding aggregation and substituted with \(E^{di}\) for distractor embedding aggregation, same as the notations used in Eq. \ref{eq:refine}. The \(update()\) follows the GRU state calculation in \cite{Cho2014}, while dot products are replaced by convolutional layers to allow two-dimensional inputs.

\paragraph{Joint learning of memory aggregation and embedding extraction.}
Target reference embedding and distractor reference embedding retrieved from respective memories are required to be well separated in their embedding space. Otherwise, outputs of attention modules would not produce correct attentions on designated locations, but generates similar level of responses across regions belonging to any object instead. Meanwhile, since embeddings used for data association are originally required to be distinguishable without aggregation, we expect all embeddings to be discriminative to identities both before and after memory aggregation.
To this end, we optimize the temporal aggregation and embedding formation with a joint discriminative learning process. An embedding of a target extracted from backbone network output firstly goes through four state initialization convolution layers, which make it distinguishable among different identities. If it is not the first embedding in a sequence, then we update it in the way described in Eq. \ref{eq:memory} for further aggregation and learn discriminability as well. Both processes are jointly trained with two discriminative identity losses, one cross-entropy loss computed between predicted identities and ground truth identities, and another triplet loss to maximize inter-identity difference and minimize intra-identity distance.

When we feed an input of length 1 for aggregation, we obtain a resultant embedding without actual aggregation. As such, we can use the memory module as an embedding extraction approach in our framework by feeding single-length inputs. With longer input of same target, the memory aggregation continuously generates aggregated discriminative embeddings. In this way, we achieve joint optimization of memory aggregation and embedding extraction.


\paragraph{Synergy between tasks.}
With aggregated discriminative embeddings, more correct attentions can be obtained in attention modules with focus on targets and suppression on distractions. Applying the attentions in position prediction lead to stronger resistance to drifts. Conversely, with more correct and prolonged predicted bounding boxes in complex scenes, we can accumulate more embeddings which are refined by attention modules, and feed them for more reliable representation through the aggregation. In this regard, temporal-aware target attention and distractor attention described in Sec. \ref{sec:tada} and identity-aware temporal aggregation presented in Sec. \ref{sec:aggregation} closely work together to form a link between the task of position prediction and the task of embedding association, thereby bring a cross-task synergy.

\section{Experiments}
The tracking performance of our proposed method is evaluated on MOTChallenge benchmarks of MOT16, MOT17, and MOT20. We also conduct an ablation study with analyses to verify the effectiveness of our design.

\paragraph{MOTChallenge and metrics.}
The MOTChallenge provides benchmarks for comparing performance of different multi-object tracking algorithms. It contains multiple pedestrian tracking scenes with various conditions like lighting, crowdness, and camera motion. The most commonly accepted challenges for benchmark are MOT16 and MOT17, both consisting of 7 video sequences for training and 7 sequences for testing. All sequences are provided with public detections. MOT17 comes with detections from three different object detectors, DPM \cite{Felzenszwalb2010}, Faster R-CNN \cite{Ren2015}, and SDP \cite{Yang2016}. MOT16 contains the same sequences, but only has DPM as public detection source and its ground truth boxes for training are less accurate than MOT17. A newer benchmark MOT20 aims to test performance under extremely crowded scenarios, which contains 4 sequences for training and 4 sequences for testing. Compared to MOT16 and MOT17, MOT20 is not yet tested by many methods due to its late emergence, but it can still provide insights on directions in complex scenes. Performance of a tracker is evaluated from several aspects by a number of metrics, while the main factors are Multiple Object Tracking Accuracy (MOTA) and ID F1 score (IDF1). MOTA measures overall performance of a tracker by evaluating errors from three sources, namely False Negatives (FN), False Positives (FP) and Identity Switches (IDS) \cite{Milan2016a}. IDF1 focuses on the quality of assigned identities on detections with a uniformed scale \cite{Ristani2016}. 
To make fair comparisons among trackers, all tracking experiments are based on the public detections provided by the MOTChallenge. Particularly, object trajectories are only initiated after the first time they appear in provided public detections.

\paragraph{Implementation details}
Experiments are conducted on a desktop with RTX 2080 Ti GPU using PyTorch. We pretrain our backbone of ResNet101 \cite{He2016} parameters on COCO dataset \cite{Lin2014}, then train on respective MOT dataset with all ground truth labeled objects with a minimum visibility of 0.1. The RPN anchor ratios are set to \(\{1.0, 2.0, 3.0\}\).
We sample 2 image frames per batch and pick 256 proposals in each image from all RPN proposals with a positive proposal sampling ratio of 0.75.
We warm up the training with memory module for embedding and aggregation learning for 3 epochs to achieve faster convergence, where learning rate is set to 0.2. We then jointly train all components for 12 epochs with an initial learning rate of 0.002 that decays by 0.5 in every 3 epochs. \(o_{min}\) is set to 0.2 empirically.
It takes around 9 hours for training on MOT16 and MOT17 to finish on a single GPU, and 15 hours on MOT20 with two GPUs.

\begin{table}[t]
\centering
\small
\setlength\tabcolsep{2pt}
\begin{tabular} {c c c c c c c c c c c c c c c c c c}
	\hline
	\hline
	& Method & MOTA\(\uparrow\) & IDF1\(\uparrow\) & FP\(\downarrow\) & FN\(\downarrow\) & IDS\(\downarrow\)\\
	\hline\hline
   \multirow{6}{*}{\rotatebox[origin=c]{90}{MOT16}}
        & MOTDT16\cite{Chen2018} & 47.6 & 50.9 & 9253 & 85431 & 792 \\
        & KCF16\cite{Chu2019a} & 48.8 & 47.2 & 5875 & 86567	& 906 \\
    	& DeepMOT\cite{Xu2020a} & 54.8 & 53.4 & 2955 & 78765 & 645 \\
    	& Tracktor++V2\cite{Bergmann2019a} & 56.2 & 54.9 & \textbf{2394} & 76844 & 617 \\
    	& GSM\cite{Liu2020} & 57.0 & 58.2 & 4332 & 73573 & \textbf{475} \\
    	& TADAM(ours) & \textbf{59.1} & \textbf{59.5} & 2540 & \textbf{71542} & 529 \\
	\hline
	\hline
	\multirow{7}{*}{\rotatebox[origin=c]{90}{MOT17}}
        & MOTDT17\cite{Chen2018} & 50.9 & 52.7 & 24069 & 250768 & 2474 \\
    	& FAMNET\cite{Chu2019} & 52.0 & 48.7 & 14138 & 253616 & 3072 \\
        & UMA\cite{Yin2020} & 53.1 & 54.4 & 22893 & 239534 & 2251 \\
    	& DeepMOT\cite{Xu2020a} & 53.7 & 53.8 & 11731 & 247447 & 1947 \\
    	& Tracktor++V2\cite{Bergmann2019a} & 56.3 & 55.1 & \textbf{8866} & 235449 & 1987 \\
    	& GSM\cite{Liu2020} & 56.4 & 57.8 & 14379 & 230174 & \textbf{1485} \\
    	& TADAM(ours) & \textbf{59.7} & \textbf{58.7} & 9676 & \textbf{216029} & 1930 \\
    \hline
	\hline
    \multirow{3}{*}{\rotatebox[origin=c]{90}{MOT20}}
        & SORT20\cite{Bewley2016} & 42.7 & 45.1 & 27521 & 264694 & 4470 \\
        & Tracktor++V2\cite{Bergmann2019a} & 52.6 & \textbf{52.7} & \textbf{6930} & 236680 & \textbf{1648}\\
	   	& TADAM(ours) & \textbf{56.6} & 51.6 & 39407 & \textbf{182520} & 2690 \\
	\hline
\end{tabular}
\caption{Comparison with modern online methods on provided public detections of MOTChallenge benchmarks. Best result in each metric is marked in \textbf{bold}.}
\label{tab:main result}
\end{table}

\subsection{Benchmark Evaluation}
The performance of our tracker is evaluated on the test set of MOTChallenge benchmarks. For MOT17, the final result is computed on all three provided public detection sets. We compare against modern online method that are officially published on the benchmark with peer-reviews.

As shown in Table \ref{tab:main result}, the benchmark results show the superior performance of our method over other published online public methods on MOT16, MOT17, and MOT20. It is noteworthy that we achieve best IDF1 and FN, in addition to the highest MOTA, on MOT16 and MOT17. Since we employ two attention modules to assist position prediction, one for enhancing focus on targets themselves and another for reducing distractions from neighbors, identities of objects are well taken care of, therefore it is not surprising to see that our proposed method performs well in terms of identity correctness indicated by IDF1. Although IDS of our method is not the best, ranking second on MOT16 and third on MOT17, it performs sufficiently well considering the minimal design of our embedding extraction. Methods with better IDS adopt more complicated models, like GSM \cite{Liu2020} trained a graph similarity model. Compared with DeepMOT \cite{Xu2020a} that also employed an integrated identity embedding optimization, we have lower IDS on both MOT16 and MOT17 with much higher IDF1 on both datasets. On the other hand, our tracker has more FP than Tracktor++V2 \cite{Bergmann2019a}, which indicates our enhanced position prediction is not always correct and could have produced FP in the process. Still, decrease in FN exceeds rise in FP by far compared with their result in both MOT16 and MOT17, which implies the effectiveness of our design.
For MOT20, our method presents a best result among published online methods, where FN is notably less than other methods.
To sum up, the benchmark results demonstrate our tracker's strong performance, and detailed analysis of it is conducted through an ablation study in Sec. \ref{sec:ablation}.

\begin{table}[t]
\centering
\small
\setlength\tabcolsep{2pt}
\begin{tabular} {l c c c c c c c c c c c c c c c c c}
	\hline
	\hline
	Setup & MOTA\(\uparrow\) & IDF1\(\uparrow\) & FP\(\downarrow\) & FN\(\downarrow\) & IDS\(\downarrow\)\\
	\hline\hline
	w/o TA \& DA & 65.9 & 71.1 & 597 & 37501 & 208 \\
	w/o DA & 66.4 & 71.2 & \textbf{462} & 37060 & 191 \\
	w/o TA & 66.7 & 71.3 & 473 & 36748 & \textbf{188} \\
	w/o adaptive weight & 66.0 & 68.5 & 679 & 37322 & 242 \\
	w/o memory aggregation & 66.5 & 67.5 & 552 & 36848 & 232 \\
	Full model & \textbf{67.0} & \textbf{71.6} & 583 & \textbf{36287} & 197 \\
	\hline
\end{tabular}
\caption{Ablation study on components on FRCNN of MOT17 train set. "TA" stands for target-awareness, "DA" for distractor-awareness. Best result in each metric is marked in \textbf{bold}}
\label{tab:ablation study: components}
\end{table}

\subsection{Ablation Study}
\label{sec:ablation}
An ablation study is conducted on MOT17 training set with provided FRCNN public detections. As shown in Table \ref{tab:ablation study: components}, we remove proposed components to see their contributions to our method. With both TA and DA modules removed, the tracking performance measured by MOTA decreases by 1.1, with worse result in all other metrics. The difference in performance mainly comes from the number of FN, where FN in full model significantly reduces, while FP also slightly decreases. This shows that more correct predictions are made with TA and DA enabled. Meanwhile, better IDF1 and fewer IDS indicate that the full model performs better in distinguishing identities with TA and DA. We see better prediction results with attention modules, which shows the benefit of adopting attentions from memory aggregated embeddings in prediction. Meanwhile, embeddings used for association are also improved, as demonstrated by the stronger discriminability. The higher performance in both tasks confirms the synergy in between.

Compared with the case where no TA and DA is employed, introducing TA without DA improves MOTA by 0.5, while applying DA alone leads to 0.8 higher MOTA. Specifically, least FP is seen in TA only setup, while lowest IDS is observed with only DA enabled. They both can improve tracking performance on their own and produce attention to bring synergy. However, the improvement is not as large as having them work together.

\begin{figure}[t]
\begin{center}
    \includegraphics[width=1.0\linewidth]{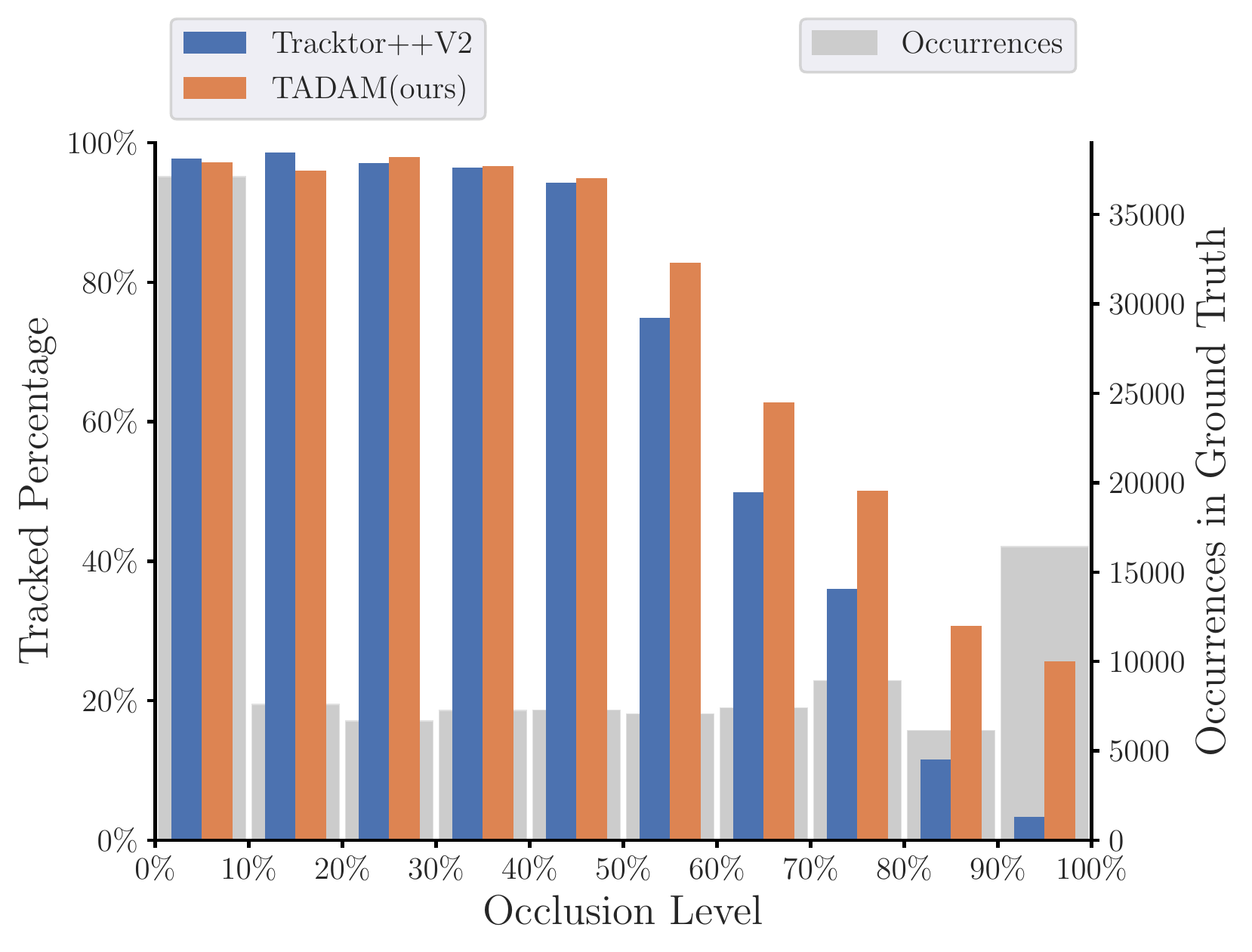}
\end{center}
   \caption{Occlusion level vs tracked percentage. Wider gray bars show the occurrence of ground truth object bounding boxes in each occlusion level interval, while narrower colored bars illustrate the percentage of objects tracked within each interval for their respective method. Note that occurrences and tracked percentages are not drawn in same unit.}
\label{fig:visibility}
\end{figure}

\paragraph{Effect of adaptive weight.}
To verify the necessity of adaptive weight, we treat all cases in the same way by removing adaptive weight regardless of occlusion levels. A drop of 1.0 in MOTA is observed comparing with the full model, which makes the performance even worse than running with TA or DA alone, though still slightly higher than that without both TA and DA. This implies that naively applying TA and DA for all cases is indeed a suboptimal approach. While occlusions have to be taken care of, it can be seen from Fig. \ref{fig:visibility} that easy cases with slight occlusions dominates the dataset, where prediction without attention modules works sufficiently well. Therefore, it is necessary to leave them untouched and apply stronger attention only for harder cases with severer occlusions.

\paragraph{Effect of memory aggregation.}
We also conduct an experiment to see the benefit of our memory aggregation. Instead of using memory aggregation, we only store target and distractor embeddings extracted in previous frame as references for TA and DA. Without the aggregation, we observe a decrease in MOTA of 0.5, as well as worse result in FP, FN, IDF1, and IDS. This indicates that the discriminative memory aggregation significantly helps the TA and DA to form robust attentions for both prediction and embedding, therefore leads to stronger performance in tracking.



\paragraph{Performance in different occlusion levels.}

An intuitive way to verify the robustness of a tracker to occlusions is to check how many of the occluded targets are being tracked under different occlusion levels. In the annotation of MOT17 train set, we have access to the levels of occlusion of all ground truth objects. With occlusion levels quantized into intervals of 10\%, occurrence of occlusion degrees within each interval can be counted to show the distribution of object visibility. Meanwhile, by calculating the percentage of ground truth objects that are tracked in each interval, we can evaluate a tracker's performance under different occlusion levels.
To determine if a ground truth bounding box is covered by a tracker, its IoU is computed with all tracked targets' boxes in the same frame and compared with a threshold of 0.5 \cite{Ristani2016}.
We compare our tracking result with Tracktor++V2 \cite{Bergmann2019a} as shown Fig. \ref{fig:visibility}.

It is observed that our tracker has very similar tracked percentage with Tracktor++V2 \cite{Bergmann2019a} when object occlusion level is less than 50\%. This implies that such position prediction with visual cues has achieved solid performance under low to medium levels of occlusion and leave little room for improvement. For objects with more than 50\% occluded, our framework shows its advantage. The higher the occlusion level, the larger the performance boost with our tracker. This is highlighted on occlusion level $>$90\%, where our tracker achieves around 25\% tracked ratio against approximately 5\%. This experiment confirms that our tracker does have better performance when dealing with occlusions. Nevertheless, how to track better with extremely low visibility could still be a direction in future research.

\section{Conclusion}
In this paper, we have proposed a method that jointly optimizes position prediction and embedding association with mutual benefits.
The two tasks are bridged by a target attention module and a distractor attention module, as well as an identity-aware memory aggregation.
The designed attention modules strengthen prediction by forcing more attention on targets and less interference from distractors, which enables extraction from more reliable embeddings for the association. On the other hand, these embeddings are exploited to form attention in prediction with the help of the memory aggregation module, and therefore assist in suppressing drifts. In this way, a synergy between the two tasks has been formed, which shows strong robustness in complex scenarios with heavy occlusions.
In our experiments, we have demonstrated the remarkable performance of our method and the effectiveness of proposed components with extensive analyses. We expect that our method can pave the way for future research to reveal potential cross-task benefits in multi-task problems like MOT.


\section*{Acknowledgement}
This work was supported in part by Australian Research Council Projects under Grant FL-170100117, Grant DP-180103424 and Grant IH-180100002, Grant IC-190100031.

{\small
\bibliographystyle{ieee_fullname}
\bibliography{mot_camera_ready}
}

\end{document}